\documentclass[runningheads]{llncs}
\usepackage[T1]{fontenc}
\usepackage{graphicx}
\usepackage{orcidlink}
\usepackage{bbding}
\usepackage{booktabs}
\usepackage{multirow}

\usepackage{amsmath}
\usepackage{pifont}
\usepackage{amssymb}

\begin{document}
\title{ShipTraj-R1: Reinforcing Ship Trajectory Prediction in Large Language Models via Group Relative Policy Optimization}

\titlerunning{ShipTraj-R1}

\author{Yang Zhan\inst{1,2}\orcidlink{0000-0001-7849-634X} \and Yunhao Li\inst{3} \and Chao Zhang\inst{3} \and Yuxu Lu\inst{2}\orcidlink{0000-0001-9845-7516}  \and Yan Li\inst{3}\orcidlink{0000-0002-4025-5070}~\Envelope
}
\authorrunning{Y. Zhan et al.}
\institute{
School of Artificial Intelligence, Optics, and Electronics (iOPEN), \\ Northwestern Polytechnical University, Xi'an 710072, China \\
\and
Department of Logistics and Maritime Studies, \\ The Hong Kong Polytechnic University, Hong Kong 999077, China \\
\and
State Key Laboratory of Information Engineering in Surveying, Mapping and Remote Sensing, Wuhan University, Wuhan 430079, China \\
\email{zhanyangnwpu@gmail.com, liyanWHU@whu.edu.cn} 
}

\maketitle              
\begin{abstract}
Recent advancements in reinforcement fine-tuning have significantly improved the reasoning ability of large language models (LLMs). In particular, methods such as group relative policy optimization (GRPO) have demonstrated strong capabilities across various fields. However, applying LLMs to ship trajectory prediction remains largely unexplored. In this paper, we propose \textbf{ShipTraj-R1}, a novel LLM-based framework that reformulates ship trajectory prediction as a text-to-text generation problem. (1) We design a dynamic prompt containing trajectory information about conflicting ships to guide the model to achieve adaptive chain-of-thought (CoT) reasoning. (2) We introduce a comprehensive rule-based reward mechanism to incentivize the reasoning format and prediction accuracy of the model. (3) Our ShipTraj-R1 is reinforced through the GRPO mechanism guided by domain-specific prompts and rewards, and utilizes the Qwen3 as the model backbone. Extensive experimental results on two complex and real-world maritime datasets show that the proposed ShipTraj-R1 achieves the least error compared with state-of-the-art deep learning and LLM-based baselines.

\end{abstract}

\section{Introduction}
\label{sec:introduction}
The surge in global maritime trade and the increasing density of ship traffic in complex waterways have intensified the demand for accurate and robust ship trajectory prediction \cite{li2024geohash}. Accurate prediction is critical for advancing maritime intelligent transportation services and ensuring navigation safety through effective collision avoidance and congestion management \cite{liang2024estimation,zhao2025diffsctg}.
Traditional methods primarily rely on Deep Learning (DL) models, such as Long Short-Term Memory (LSTM) and Convolutional Neural Networks (CNNs), and their hybrids, demonstrating a strong ability to capture sequential and spatio-temporal features.

Recently, Large Language Models (LLMs) has shown great potential in sequence generation and complex reasoning tasks across various domains \cite{grattafiori2024llama,li2025uptm}. Leveraging their powerful semantic understanding and inherent reasoning capabilities, researchers have begun exploring LLMs for trajectory prediction. Approaches like LMTraj-SUP \cite{bae2024can} and LG-Traj \cite{chib2025lg} encode continuous coordinates into symbolic tokens, demonstrating the potential of LLMs to generate trajectories autoregressively, analogous to generating natural language sentences.

However, the application of LLMs specifically for ship trajectory prediction remains largely unexplored. Crucially, even the existing LLM-based prediction models are black box regressors that rely on supervised fine-tuning. These methods imitate historical trajectories, which is insufficient. The cost of error in collision avoidance is extremely high, requiring that the complex reasoning process of the prediction model not only refer to historical data but also take into account the status of the surrounding ships, especially those with collision risks.

To address these challenges, we propose ShipTraj-R1, a novel LLM-based prediction framework built upon reinforcement fine-tuning and Chain-of-Thought (CoT) reasoning. Our method fundamentally reformulates ship trajectory prediction as a text-to-text generation problem. The core of ShipTraj-R1 lies in its ability to enforce domain-specific reasoning and precision by employing the Group Relative Policy Optimization (GRPO) mechanism. 
Specifically, we design a dynamic prompt incorporating trajectory information about conflicting ships to induce adaptive CoT reasoning, which is essential for proactive collision avoidance.
Then, the reinforcement process is dynamically guided by a novel rule-based reward function that simultaneously incentivizes the model’s structured reasoning format and quantifies coordinate prediction accuracy, ensuring reliability in complex maritime scenarios. 
Finally, by fine-tuning our model, ShipTraj-R1 can iteratively self-improve its reasoning and prediction capabilities in complex scenarios.
To the best of our knowledge, we have for the first time provided a benchmark evaluation of LLMs in the ship trajectory prediction.

Our contributions can be summarized as follows:
\begin{itemize}
\item We propose ShipTraj-R1, a novel LLM-based ship trajectory prediction framework. It reformulates the traditional numerical regression task into a text-to-text generation problem with CoT reasoning.
\item We design a prompt containing context about conflicting ships, rule-based reward mechanisms and the reinforcement fine-tuning paradigm using GRPO, specifically for ship trajectory prediction.
\item We provide a comprehensive benchmark on real-world ship trajectory data in various representative and SOTA models and LLM-based baselines.
Extensive experiments demonstrate that ShipTraj-R1 is promising and of great potential for ship trajectory prediction.
\end{itemize}

\section{Related Work}
\label{sec:relatedwork}

\subsection{Ship Trajectory Prediction}
Deep learning (DL) methods have become highly competitive in ship trajectory prediction \cite{li2024geohash} and are mainly divided into three categories: discriminative, generative, and hybrid methods.
Discriminative methods are widely used, learning direct mappings from input to output. LSTM and Gate Recurrent Unit (GRU) can effectively capture long-term dependencies in trajectory sequences and have become mainstream. CNNs are often combined with these temporal models \cite{lu2021novel} to capture comprehensive spatiotemporal information.
Furthermore, graph-based models like Graph Convolutional Neural Networks (GCNN) and Graph Auto-Encoders (GAE), are applied to effectively model the mutual influence and interactive behaviors among individual ships \cite{wang2024vessel}.
Generative methods are designed to comprehend data distributions and produce new data samples. This category includes Generative Adversarial Networks (GANs) \cite{liu2021lstm}, with specific variants like Time-series GAN and Sequence GAN \cite{yu2017seqgan} being highly relevant for time-series analysis.
Hybrid methods provide an effective and adaptable approach by combining two or more DL models.

\subsection{LLM-based Trajectory Prediction}
The application of LLMs for trajectory prediction is an emerging research direction, rapidly expanding due to their powerful semantic and reasoning capabilities \cite{zhan2025skyeyegpt}.
Existing methods can be broadly divided into direct language-based modeling and LLM-based auxiliary prediction.
For direct language-based modeling, trajectory data is adapted into textual format for LLMs, enabling them to predict future trajectory coordinate "tokens" akin to generating words.
For example, LMTraj-SUP \cite{bae2024can} reformulates the task as question-answering. It converts numerical and scene features into structured prompts and utilizes a decimal-aware tokenizer to bridge the gap between continuous values and discrete tokens. Traj-LLM \cite{lan2024traj} integrates a frozen GPT-2 backbone directly into the prediction pipeline. It encodes multimodal scene context into transformer-compatible token sequences and employs LoRA for fine-tuning, demonstrating strong few-shot adaptability.
LLMs also serve several critical auxiliary functions in this domain. 
Language-guided scene understanding uses language as a complementary modality to enrich spatial context \cite{chib2025lg}. By encoding semantics like scene images or traffic rules, models achieve more interpretable and socially aware predictions.
Language-driven data generation leverages LLMs as generative engines to synthesize diverse and controllable trajectory datasets from high-level textual descriptions \cite{yang2025trajectory}. 
Finally, language-based reasoning and interpretability employs LLMs to explain or justify a model's predictions \cite{peng2025lc}. This often involves CoT prompting to verbalize the model's decision-making process, enhancing safety and user trust.
However, despite this significant progress, the application of LLMs for ship trajectory prediction remains a largely unexplored area.

\section{Methodology}
\label{sec:methods}

\subsection{Problem Definition}
\label{sec:prob_defi}

\begin{figure}[t]
\centering.
\vspace{-8pt}
\includegraphics[width=\textwidth]{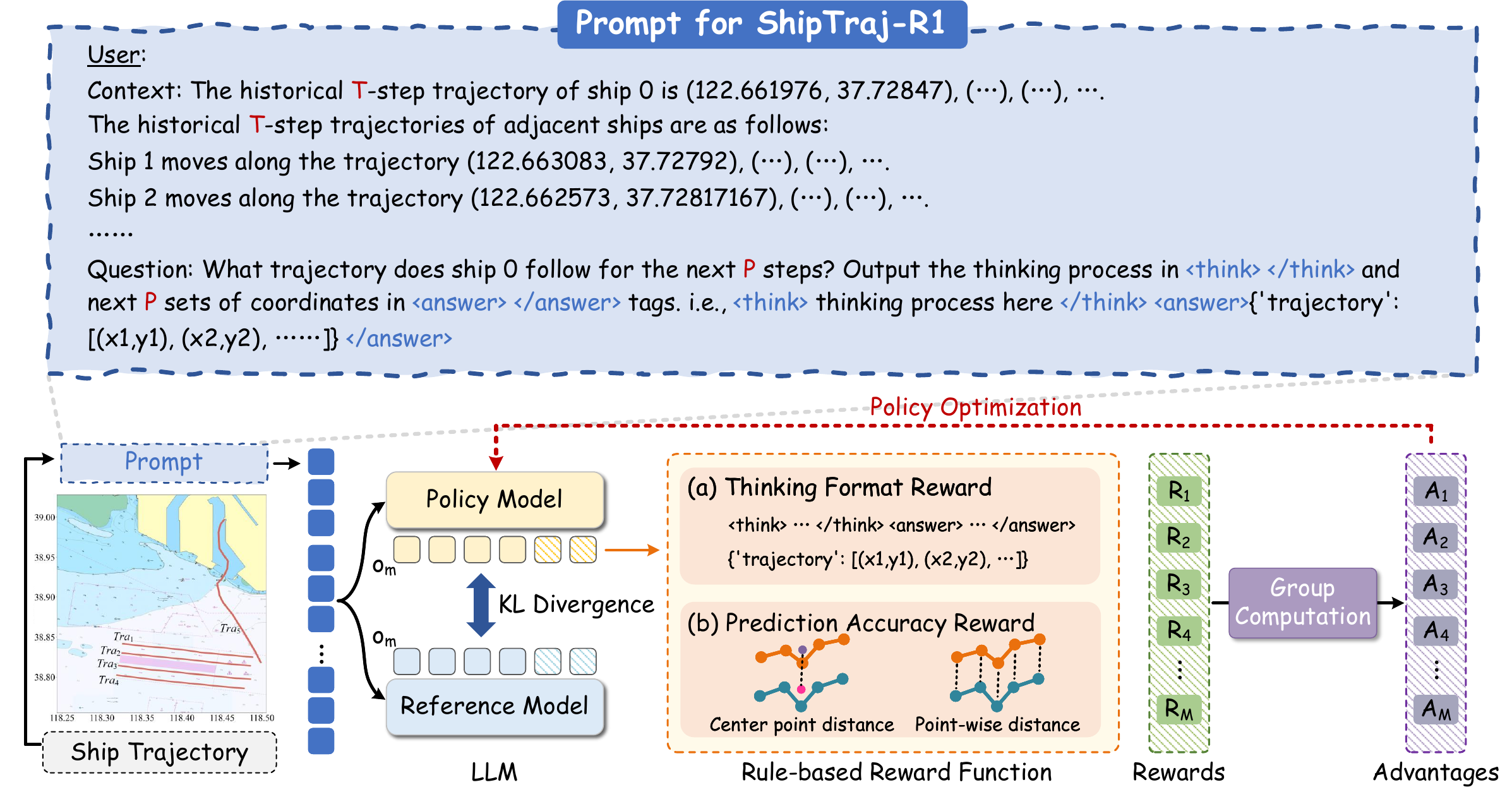}
\caption{Overview of our proposed ShipTraj-R1 framework.}
 \vspace{-12pt}
\label{model}
\end{figure}

The objective of ship trajectory prediction is to forecast a ship's future coordinates given a sequence of its past observations. 
Formally, the historical trajectory $\mathbf{X}_i$ with length $T_{obs}$ for ship $i$ can be defined as a sequence of spatial points 
$\mathbf{X}_i = \{(lon_i^t, lat_i^t) | t \in [1, \dots, T_{obs}]\}$, where $(lon_i^t, lat_i^t)$ is the longitude and latitude at time step $t$.
The task is to predict the ground truth future trajectory $\mathbf{Y}_i = \{(lon_i^t, lat_i^t) | t \in [T_{obs}+1, \dots, T_{obs}+T_{pred}]\}$.
The prediction model $\mathcal{F}$ seeks to minimize the error or distance between the predicted trajectory $\mathbf{\hat{Y}}_i = \mathcal{F}(\mathbf{X}_i)$ and the ground truth trajectory $\mathbf{Y}_i$.

\subsection{ShipTraj-R1 Model}
\label{sec:model}
We propose ShipTraj-R1, a specialist LLM for ship trajectory prediction. The overall architecture of ShipTraj-R1 is illustrated in Figure \ref{model}.
The framework is built upon two core principles: (1) reformulating the numerical prediction into a text-to-text generation task, and (2) leveraging a powerful, open-source LLM model as the core reasoning agent.

\noindent \textbf{Task Reformulation and Model}
As defined in Section \ref{sec:prob_defi}, existing trajectory prediction is treated as a numerical regression problem. They typically encode continuous GPS coordinates through normalization or specialized schemes, use models with a decoder for numerical prediction, and then apply inverse transformations.
In contrast, we instead adopt a "first-reasoning, then-predicting" paradigm. As an LLM-based approach, ShipTraj-R1 reformulates the task as a text-to-text generation problem with CoT reasoning.
The input observation $\mathbf{X}_i$ is textualized and attached to a structured prompt $P$. The model then generates a structured output $\mathbf{O}$ that fuses the explicit CoT reasoning $think$ and the textualized predicted trajectory $\mathbf{\hat{Y}}_i$:
\begin{equation}
  think, \; \mathbf{\hat{Y}}_i =  \mathcal{F}_{ShipTraj-R1}\;(\mathbf{X}_i, \; P) \; .
\end{equation}
The text-based nature eliminates the need for normalization/inverse-normalization steps of longitude and latitude, simplifying the model architecture. The explicit reasoning provides an intermediate interpretability process that justifies its prediction.
ShipTraj-R1 is built upon the Qwen3 model \cite{yang2025qwen3}, which serves as the foundational reasoning and prediction backbone.
Qwen3 serves as our foundation due to its demonstrated superior performance in mathematical reasoning and structured text generation across various benchmarks, making it highly suitable for the coordinate-level precision required in trajectory prediction.

\noindent \textbf{Prompt Design Contextualizing Conflicting Ships}
We design a dynamic prompt that explicitly incorporates the contextual information of the trajectories of nearby ships with a significant risk of conflict. In maritime navigation, this is crucial because the trajectory is not purely kinematic but is dominated by collision avoidance logic.
As illustrated in Figure \ref{model}, the prompt serves as a cognitive scaffold that guides the model through a "first-reasoning, then-predicting" paradigm. 
The prompt structure explicitly provides the necessary spatio-temporal context, including the historical trajectory of the target ship and the trajectories of adjacent ships that pose a risk of conflict.
Our methodology leverages the inherent capability of LLMs to formulate CoT reasoning without requiring labor-intensive, manually annotated reasoning data. Instead of being constrained to mimic fixed logical patterns, the model is incentivized to autonomously generate an adaptive thinking process that is tailored to the adjacent ship context of each trajectory sample.
This comprehensive context is critical for prompting the LLM to perform adaptive social-spatial reasoning.

\textbf{Ship Domain Conflict Detection Algorithm:}
We establish a quaternion ship domain (QSD) model to conduct ship domain conflict detection \cite{wang2010intelligent}.
This step assesses the collision risk between the raw adjacent ship and the target ship.
The trajectory information of conflicting ships that exceed the risk threshold is provided to the model instead of providing trajectories of all neighboring ships.
The QSD model defines a ship's requisite safe water domain, which other ships must avoid to prevent collision risk.
The QSD is defined in the ship-centered coordinate system by the four safety distance: $R_{bow}$ (bow), $R_{stern}$ (stern), $R_{starb}$ (starboard), and $R_{port}$ (port). The boundary of the QSD is defined as:
\begin{equation}
\text{QSD} = \{(x, y) | f_k(x, y; H) \le 1, H=\{R_{fore}, R_{aft}, R_{starb}, R_{port}\}\} ,
\end{equation}
where $k$ determines the shape of the QSD. When $k$=1, its shape is a combined quadrilateral. When $k$=2, its shape is a combined ellipse.

\subsection{Rule-based Reward Function}
\label{sec:reward}
To reinforce the LLM’s ability to generate both reliable CoT reasoning and accurate coordinate outputs, we design a comprehensive rule-based reward function, including the thinking format reward and the prediction accuracy reward. 

\noindent \textbf{Thinking Format Reward}
This reward ensures the model's output is correctly structured and reliably parsable for accuracy evaluation. The reward is binary ($1$ or $0$) for compliance.
The model's response must strictly adhere to the required structure, separating the reasoning and the final answer using explicit tags, \textit{i.e.}, \textit{<think> the CoT thinking </think>} \textit{<answer> $\dots$ </answer>}.
Within the $\langle \text{answer} \rangle \langle /\text{answer} \rangle$ tag, the contained content must be a valid trajectory object. This object must contain the key "trajectory" whose value is a list of exactly $T_{pred}$ predicted coordinate pairs. 
Furthermore, each coordinate pair must fall within the predefined geographic boundary constraints of the maritime region. Under the above constraints, the trajectory format is considered correct.

\noindent \textbf{Prediction Accuracy Reward}
This task-specific reward quantifies the error between the predicted trajectory $\mathbf{\hat{Y}}_i$ and the ground truth trajectory $\mathbf{Y}_i$. We use the Vincenty formula to calculate the geodesic distance between coordinates.
First, we calculate the center point of both the ground-truth and predicted trajectories by averaging their respective coordinates, \textit{i.e.}, $\mathbf{C}_{i}$ and $\mathbf{\hat{C}}_{i}$. 
A reward of 1 is assigned if their distance is less than 120 meters; otherwise, the reward is 0.
Then, we need to encourage each predicted trajectory point to achieve high precision.
We calculate the distance between the predicted coordinates and the true coordinates for each point. For each distance below the threshold of 90 meters, we increment the reward by $\frac{1}{T_{pred}}$.

\subsection{Reinforcement Fine-Tuning}
\label{sec:train}

The ultimate objective of ShipTraj-R1 is to elevate the LLM's ability to produce highly accurate and safety-compliant ship trajectory predictions. We achieve this by optimizing the model using the GRPO algorithm \cite{shao2024deepseekmath}, guided by our designed rule-based rewards.
For each input $x$, which includes the historical trajectory and adjacent ship context, the old policy model $\pi_{\theta_{old}}$ samples $M$ candidate output completions (CoT reasoning and trajectories) $\{o_m\}_{m=1}^{M}$.
Each completion $o_m$ is evaluated by the rule-based reward function to obtain a scalar reward $R_m$. This reward quantifies both the correctness of the reasoning structure and the accuracy of the predicted trajectory.
The advantage $\hat{A}_m$ for each output $o_m$ is calculated relative to the average reward of its group:
\begin{equation}
  \hat{A}_{m}=\big({R_m - \text{mean}\{R_1, R_2, \dots, R_M\}}\big)/{\text{std}\{R_1, R_2, \dots, R_M\}}.
\end{equation}
This relative advantage calculation naturally promotes samples that exhibit superior prediction accuracy than the group average, effectively utilizing feedback to steer the learning process.
The objective $\mathcal{J}_{\text{GRPO}}(\theta)$ is designed to maximize the expected advantage while ensuring the updated policy $\pi_{\theta}$ does not deviate excessively from the reference policy $\pi_{\text{ref}}$. The objective is formulated as:
\begin{equation}
\begin{split}
 \mathcal{J}_{GRPO}(\theta) 
 &= \mathbb{E}\{o_m\}_{m=1}^M \sim \pi_{\theta_{old}}(O|x)  \\
    \Big[\frac{1}{M}\!\sum_{m=1}^M&\min\!\Big(\rho^{(m)}_{\theta}\,\!\hat{A}_{m}, \text{clip} \big(\rho^{(m)}_{\theta},\,1{-}\epsilon,\,1{+}\epsilon\big)\,\!\hat{A}_{m}\Big) -  \beta\, {D}_{KL}\!\left(\pi_{\theta}\,\Vert\,\pi_{\text{ref}}\right)\Big], 
\end{split}
\label{eq:GRPO-obj}
\end{equation}
where $\rho_{\theta}^{(m)} = \frac{\pi_{\theta}(o_m|x)}{\pi_{\theta_{\text{old}}}(o_m|x)}$, $\epsilon$ is the clipping hyperparameter, and $\beta$ controls the KL-divergence regularization strength.

\section{Experiment}
\label{sec:experiments}

\subsection{Experimental Settings}

\noindent \textbf{Datasets}
To rigorously evaluate the predictive capability and stability of our proposed ShipTraj-R1 framework, we conducted extensive experiments using real-world ship trajectory data derived from the Automatic Identification System (AIS). 
Our benchmark dataset focuses on two distinct maritime regions: ChengShanJiao Promontory (CSJP) and CaoFeiDian Port (CFDP).
The waterway layouts and ship traffic patterns in CSJP waters are generally more complex than those in CFDP.
The AIS data was collected over a two-week period, from July 6 to July 19, 2020, for both areas.
Prior to model training, the raw, non-equidistant trajectory data was preprocessed using cubic spline interpolation to ensure uniform time intervals. Specifically, the trajectories were sampled at a 5-second interval per point.
The CSJP dataset contains 2,649 complete ship trajectories, including 25,592 GPS points. The CFDP dataset comprises 2,948 ship trajectories, including 34,038 GPS points.
The total dataset for each region was partitioned into 90\% for training and 5\% each for validation and testing.

\noindent \textbf{Evaluation Metrics}
We employ two widely used evaluation metrics for ship trajectory prediction: the Final Displacement Error (FDE) and the Average Displacement Error (ADE).
The FDE measures the $L_2$ distance between the predicted position and the true position solely at the final predicted time step.
The ADE measures the mean $L_2$ distance between the predicted position and the true position across all predicted time steps in a trajectory.
 Mathematically, they are defined as:
\begin{equation}
FDE = \frac{1}{M} \sum_{i=1}^{M} \sqrt{(x_{i}^{P}-\hat{x}_{i}^{P})^{2}+(y_{i}^{P}-\hat{y}_{i}^{P})^{2}},
\label{eq:FDE}
\end{equation}
\begin{equation}
ADE = \frac{1}{M \times P} \sum_{i=1}^{M} \sum_{t=1}^{P} \sqrt{(x_{i}^{t}-\hat{x}_{i}^{t})^{2}+(y_{i}^{t}-\hat{y}_{i}^{t})^{2}} ,
\label{eq:ADE}
\end{equation}
where $M$ is the number of trajectories, $P$ is the number of predicted points, $(x_{i}^{t}, y_{i}^{t})$ is the ground truth coordinate, and $(\hat{x}_{i}^{t}, \hat{y}_{i}^{t})$ is the predicted coordinate at time step $t$ for trajectory $i$. For both ADE and FDE, a lower value indicates higher prediction accuracy.

\noindent \textbf{Implementation Details}
For the GRPO training, we perform the RL-based post-training paradigm, implemented via VLM-R1 \cite{shen2025vlm}.
We adopt Qwen3-8B \cite{yang2025qwen3} as our default language model.
The initial learning rate is set to 1e-6, and the weight decay is 1e-2. 
We set the batch size to 16.
Our experiments are conducted on four NVIDIA RTX 4090 GPUs and four L20 GPUs.
For our prediction task, we utilized past observed trajectories with length $T_{obs} \in[4,8]$ (20s or 40s) to predict future trajectories with length $T_{pred} \in[1,2,3,4]$ (5s, 10s, 15s, or 20s). 
Our LLM-based model can unify prediction tasks under different $T_{obs}$ and $T_{pred}$ conditions and has strong flexibility. However, traditional DL-based methods must train separate models for different combinations of $T_{obs}$ and $T_{pred}$.

\subsection{Comparison with State-of-art Methods}

\begin{table*}[t]
\centering
\caption{
Comparisons with representative DL-based methods, LLM-based methods, and SoTA open-source LLMs.
}
\label{result_all}
\resizebox{0.95\linewidth}{!}{
\begin{tabular}{lccccccc}
\toprule
\multirow{3}{*}{\textbf{Model}} & \multirow{3}{*}{\textbf{\begin{tabular}[c]{@{}c@{}}Release\\ Year-month \end{tabular}}}  & \multicolumn{2}{c}{\textbf{CSJP}} & \multicolumn{2}{c}{\textbf{CFDP}} \\ \cmidrule(r){3-4} \cmidrule(r){5-6} 
&  & FDE & ADE &  FDE & ADE  \\ 
\midrule
\multicolumn{6}{l}{\textit{\textbf{DL-based Methods}}} \\
TBENet $_{\textcolor{gray}{\text{TRE'24}}}$ \cite{li2024bi} & 24-11  & 0.009827  & 8.1000e-04 & 0.005993	 &  4.0200e-04 \\
GeoCLSTM $_{\textcolor{gray}{\text{CAEE'24}}}$ \cite{li2024geohash} & 24-12  & 0.070173 & 5.4530e-03 & 0.064019 & 4.1890e-03 \\
DBSCAN-GeoCLSTM $_{\textcolor{gray}{\text{CAEE'24}}}$ \cite{li2024geohash} &  24-12  & 0.033494 & 2.2640e-03 & 0.030183 & 1.4730e-03 \\
GAT-LSTM $_{\textcolor{gray}{\text{JMSE'24}}}$ \cite{li2024vessel} &  24-12 & 0.012828 & 2.7260e-03 & 0.009032 &  1.0520e-03 \\
DGCN-Transformer $_{\textcolor{gray}{\text{EAAI'25}}}$ \cite{zhang2025incorporating} &  25-02 & 0.009216 & 8.2400e-04 &0.006087  &  5.8900e-04 \\
SeqLSTM-U-Net $_{\textcolor{gray}{\text{OE'25}}}$ \cite{gan2025maritime} & 25-08  & 0.009907  &  1.4360e-03 &0.007183  & 9.6300e-04 \\
\cmidrule(r){1-6}
\multicolumn{6}{l}{\textit{\textbf{LLM-based Methods}}} \\
LMTraj-SUP $_{\textcolor{gray}{\text{CVPR'24}}}$ \cite{bae2024can}&  24-03 & 0.004734  & 9.2331e-05 & 0.001297 & 5.0249e-05 \\
LG-Traj $_{\textcolor{gray}{\text{CVPR'25}}}$ \cite{chib2025lg} &  24-03 & 0.004081  &  7.9923e-05 & 0.001004 & 3.7743e-05 \\
Traj-LLM $_{\textcolor{gray}{\text{T-IV'25}}}$ \cite{lan2024traj}  & 24-06 & 0.003894 & 4.7202e-05 & 0.000674 & 9.6587e-06 \\
\cmidrule(r){1-6}
\multicolumn{6}{l}{\textit{\textbf{Open-source LLMs}}} \\
Baichuan2-7B \cite{yang2023baichuan} & 23-12 & 0.028621 & 2.1598e-04 &  0.003834 & 2.8329e-05  \\
Hunyuan-7B \cite{cao2025hunyuanimage}  & 25-07 & 0.010099  & 1.5845e-04 & 0.003294 &   1.2795e-05 \\
Llama-3-8B \cite{grattafiori2024llama} & 24-04  & 0.002671 & 7.0153e-05  & 0.000897 &  9.7625e-06 \\
Qwen2.5-3B \cite{team2024qwen2}& 24-09   & 0.003862  & 4.8339e-05 &  0.000882 &  2.8602e-06 \\
Qwen2.5-7B \cite{team2024qwen2} & 24-09   & 0.003255  & 2.9338e-05 & 0.000498 & 6.1298e-07 \\
DeepSeek-R1 \cite{guo2025deepseek}  & 25-05  & 0.002303 & 2.2688e-05  &  0.000490  &  6.9395e-07\\
Qwen3-4B \cite{yang2025qwen3} & 25-07  & 0.002350 & 2.7751e-05 &  0.000504 & 7.7631e-07 \\
Qwen3-8B \cite{yang2025qwen3} & 25-07  & 0.002293 & 2.0529e-05 & 0.000474 & 6.7403e-07 \\
\cmidrule(r){1-6}
\textbf{ShipTraj-R1-4B} (Ours) & -  & {0.001582} & {1.8084e-05}& {0.000386} & {4.9705e-07}  \\
\textbf{ShipTraj-R1-8B} (Ours) & -  & \textbf{0.001297} & \textbf{1.1547e-05}& \textbf{0.000311} & \textbf{3.8912e-07}  \\
\bottomrule                           
\end{tabular}
}
\vspace{-10pt}
\end{table*}

\noindent \textbf{Quantitative Results}
The quantitative results presented in Table \ref{result_all} establish the superior performance of our proposed ShipTraj-R1 framework across both the complex CSJP and CFDP datasets.
Firstly, our ShipTraj-R1 achieves the lowest FDE and ADE across all tested baselines, setting a new state-of-the-art for ship trajectory prediction. The results represent a marked improvement over all competitors.
The gain over traditional DL-based models is substantial. This large performance gap validates the fundamental advantage of leveraging Large Language Models over small models for complex maritime scenarios.
Furthermore, ShipTraj-R1 significantly outperforms previous SOTA LLM-based trajectory models like Traj-LLM.
On the CFDP dataset, our model reduces the FDE from Traj-LLM's $0.000674$ to $0.000311$. Crucially, by comparing ShipTraj-R1-8B against its base model, Qwen3-8B, we observe a definitive reduction in error. 
This gain directly validates the efficacy of our designed prompt and reward mechanisms in reinforcing the model's spatial reasoning based on latitude and longitude coordinates through reinforcement fine-tuning.
The consistent and pronounced error reduction across both datasets highlights that the ShipTraj-R1 approach is not only highly accurate but also robust in generalizing to diverse and challenging maritime traffic environments.


\begin{figure}[t]
\centering
\vspace{-8pt}
\includegraphics[width=0.98\textwidth]{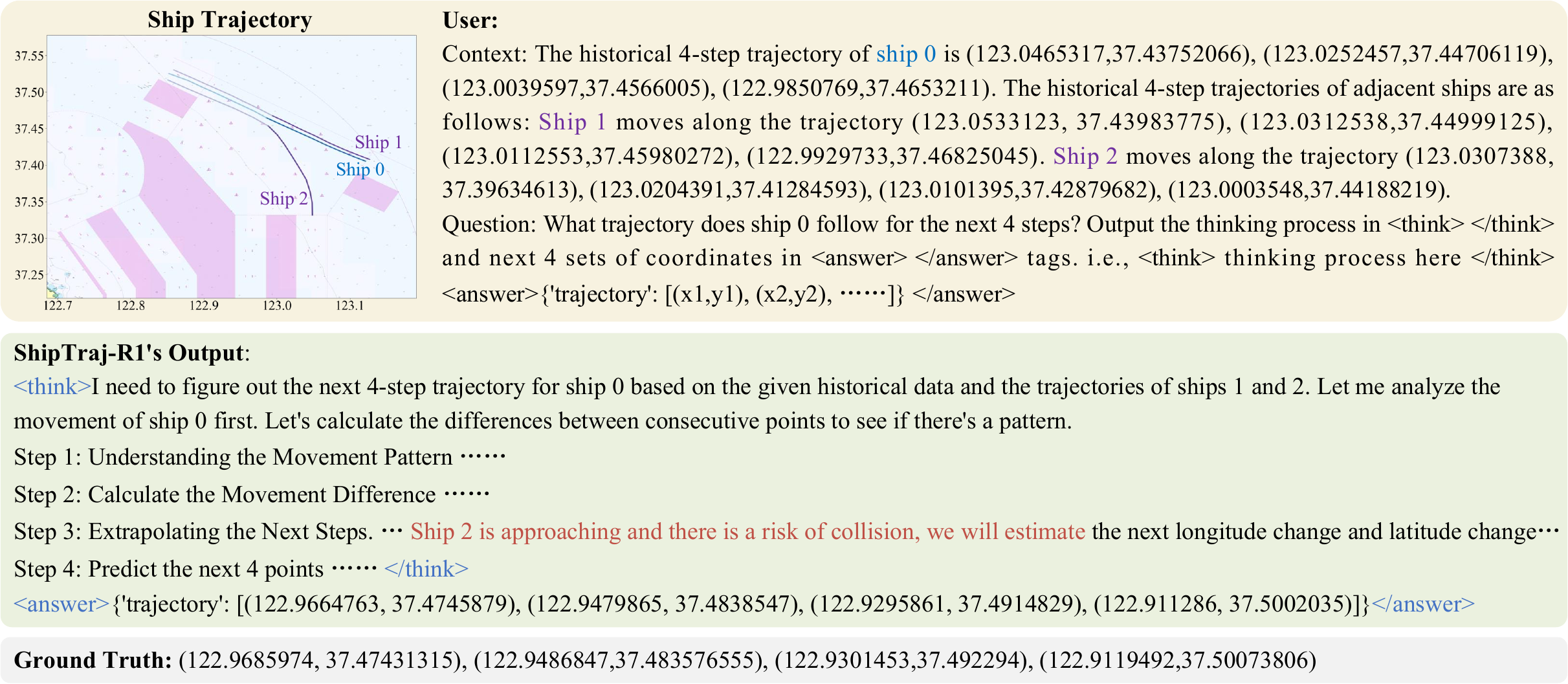}
\caption{
Qualitative examples of our ShipTraj-R1 on the CSJP dataset.
In this case, the ShipTraj-R1 adaptively generates CoT reasoning by itself based on the user prompt.
}
\vspace{-6pt}
\label{show_result2}
\end{figure}

\begin{figure}[t]
\centering
\includegraphics[width=0.8\textwidth]{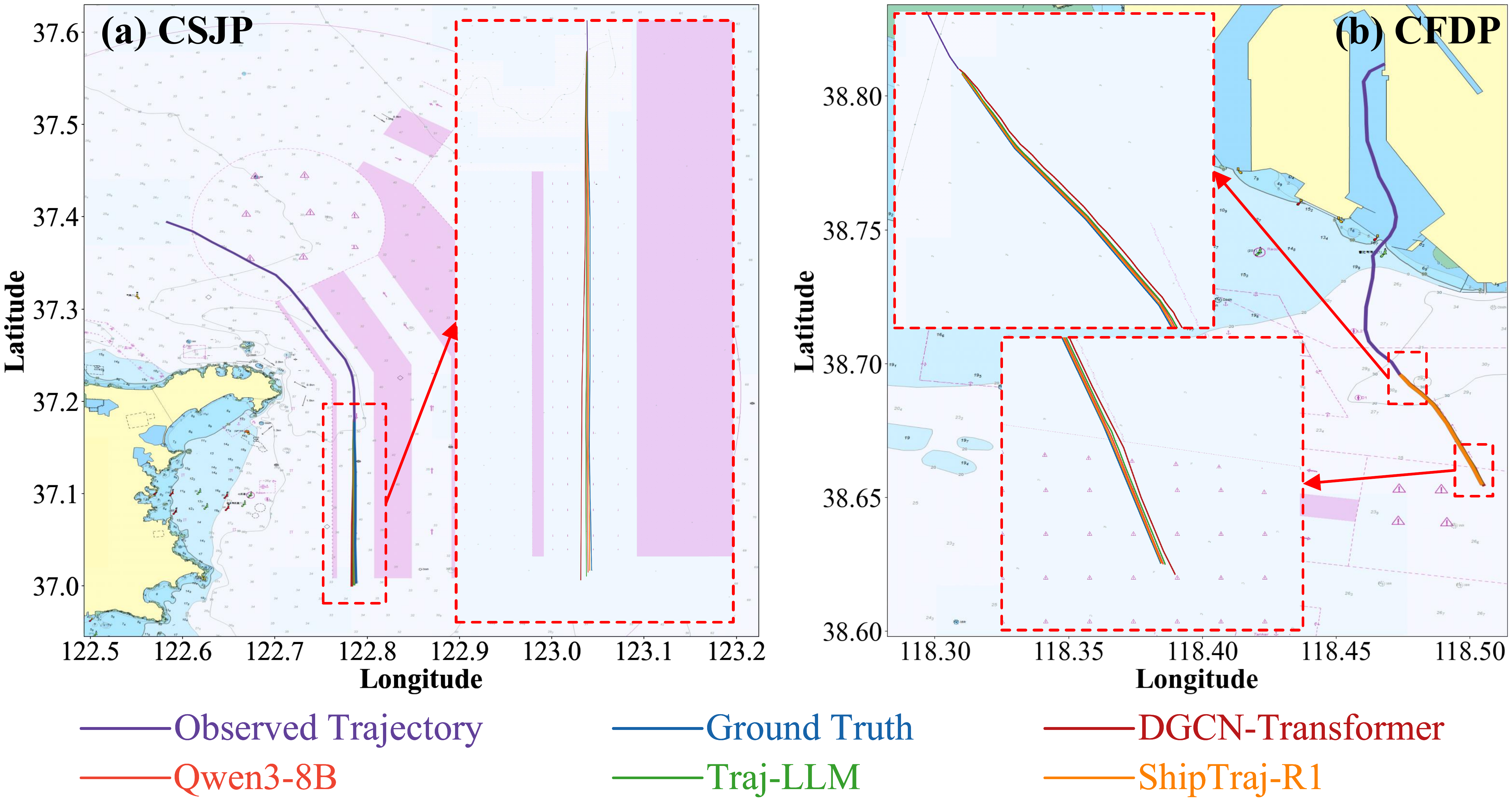}
\caption{
Illustration of predicted ship trajectories from the CSJP and CFDP datasets.
}
\vspace{-14pt}
\label{show_result}
\end{figure}

\noindent \textbf{Qualitative Results}
To visually validate the superiority of ShipTraj-R1 in complex maritime environments, we provide an example in Figure \ref{show_result2}.
We have observed that the reasoning process is helpful for physics-based motion analysis and the estimation of the impact of conflicting ships.
Figure \ref{show_result} presents comparative qualitative results against several competitive baselines across the CSJP and CFDP datasets.
DGCN-Transfromer \cite{zhang2025incorporating}, Traj-LLM \cite{lan2024traj}, and Qwen3-8B \cite{yang2025qwen3} are respectively the most advanced methods among DL-based methods, LLM-based methods, and open-source LLMs.
As seen in the zoomed-in views, ShipTraj-R1's predicted trajectory aligns almost perfectly with the Ground Truth trajectory.
The qualitative visualization confirms that ShipTraj-R1's reinforcement fine-tuning successfully addresses the limitations of pure geometric and language-based models, yielding predictions that are significantly more accurate, physically compliant, and stable, thereby minimizing the risk of collision or violation in real-world applications.

\subsection{Ablation Studies}

\begin{table*}[t]
\centering
\caption{
Ablation on different fine-tuning methods and chain-of-thought.
}
\label{sft_rl}
\resizebox{0.75\linewidth}{!}{
\begin{tabular}{lcccccc}
\toprule
\multirow{3}{*}{{Model}} & \multirow{3}{*}{{Type}} & \multirow{3}{*}{{CoT}}  & \multicolumn{2}{c}{{CSJP}} & \multicolumn{2}{c}{{CFDP}}  \\ \cmidrule(r){4-5}  \cmidrule(r){6-7} 
& & & FDE & ADE & FDE & ADE \\ 
\midrule
Baseline-8B & -  & - & 0.002293 & 2.0529e-05 & 0.000474 & 6.7403e-07 \\
ShipTraj-R1-8B & SFT & \ding{55} & {0.001701} & {1.8153e-05}& {0.000412} & {5.1295e-07} \\
ShipTraj-R1-8B & RFT & \ding{55} & {0.001430} & {1.6216e-05}& {0.000341} & {4.1216e-07} \\
ShipTraj-R1-8B & RFT & $\checkmark$ & \textbf{0.001297} & \textbf{1.1547e-05}& \textbf{0.000311} & \textbf{3.8912e-07} \\
\bottomrule                           
\end{tabular}
}
\end{table*}

\begin{table*}[t]
\centering
\caption{
Ablation on whether the user prompt includes conflicting ship context.
}
\label{prompt_result}
\resizebox{0.82\linewidth}{!}{
\begin{tabular}{lccccc}
\toprule
\multirow{3}{*}{{Model}}  & \multirow{3}{*}{{Conflicting Ships}}  & \multicolumn{2}{c}{{CSJP}} & \multicolumn{2}{c}{{CFDP}}  \\ \cmidrule(r){3-4}  \cmidrule(r){5-6} 
& & FDE & ADE & FDE & ADE \\ 
\midrule
ShipTraj-R1-8B  & \ding{55} & 0.002756 & 2.7804e-05& 0.000569 & 6.3881e-07 \\
ShipTraj-R1-8B & $\checkmark$ & \textbf{0.001297} & \textbf{1.1547e-05}& \textbf{0.000311} & \textbf{3.8912e-07} \\
\bottomrule                           
\end{tabular}
}
\vspace{-4pt}
\end{table*}

\begin{table*}[t!]
\centering
\caption{
Ablation on the the KL loss coefficient. 
}
\label{coef_loss}
\resizebox{0.8\linewidth}{!}{
\begin{tabular}{lccccc}
\toprule
\multirow{3}{*}{{Model}}  & \multirow{3}{*}{{KL loss coef}}  & \multicolumn{2}{c}{{CSJP}} & \multicolumn{2}{c}{{CFDP}}  \\ \cmidrule(r){3-4}  \cmidrule(r){5-6} 
& & FDE & ADE & FDE & ADE \\ 
\midrule
ShipTraj-R1-8B  & 1e-4 & \textbf{0.001297} & \textbf{1.1547e-05}& \textbf{0.000311} & \textbf{3.8912e-07} \\
ShipTraj-R1-8B  & 5e-4 & 0.001394 & 1.4580e-05& 0.000336 & 4.4628e-07 \\
ShipTraj-R1-8B  & 1e-3 & 0.002287 & 2.0466e-05& 0.000519 & 7.3491e-07  \\
ShipTraj-R1-8B  & 5e-3 & 0.003650 & 3.6814e-05& 0.000842 & 2.1277e-06 \\
\bottomrule                           
\end{tabular}
}
\vspace{-10pt}
\end{table*}

\textbf{SFT vs. RFT}
We compare the efficacy of our reinforcement fine-tuning (RFT) using GRPO against supervised fine-tuning (SFT), as detailed in Table \ref{sft_rl}.
The RFT-only model (without CoT) significantly outperforms the SFT model. This performance gap validates RFT’s superior ability to leverage the rule-based reward signal for enforcing prediction.
Furthermore, with CoT reasoning, our ShipTraj-R1 yields the best overall accuracy.

\noindent \textbf{User Prompt}
We investigate the impact of trajectory information about conflicting ships within the prompt, as shown in Table \ref{prompt_result}.
The omission of this conflict context results in a significant performance drop. This degradation confirms the crucial role of conflicting ship context in guiding the LLM's reasoning.

\noindent \textbf{KL Loss Coefficient}
The KL divergence regularization term balances the acquisition of new domain knowledge against the retention of the LLM’s pre-existing capabilities. As shown in Table \ref{coef_loss}, the optimal performance is achieved at the lowest coefficient 1e-4. As the coefficient increases, performance degrades.

\section{Conclusion}
\label{sec:conclusion}
In this work, we propose ShipTraj-R1, a novel framework that leverages LLMs with CoT reasoning to achieve robust ship trajectory prediction.
(1) Our method reformulates the complex prediction task as an LLM sequence generation problem and utilizes GRPO during reinforcement fine-tuning.
(2) Our model can generate an explicit CoT reasoning process to justify its predictions, enhancing interpretability.
(3) Unlike traditional methods that require separate models, our framework flexibly unifies prediction tasks under various observation ($T_{obs}$) and prediction ($T_{pred}$) lengths.
Extensive experiments on complex real-world datasets demonstrate that ShipTraj-R1 establishes a new state-of-the-art, significantly reducing both FDE and ADE compared to existing baselines.

We acknowledge that the current short-term prediction horizon (up to 20s) primarily validates the model's capability for high-precision, immediate course-keeping. However, a complete long-term collision avoidance system requires a much longer forecast window. 
We plan to explore the performance of LLMs in long-term prediction, which holds greater value for practical applications.

\subsubsection{\ackname} 
This work is supported in part by grants from the National Natural Science Foundation of China (No.624B2113).

%
%
\bibliographystyle{splncs04}
\bibliography{mybibliography}

\end{document}